\documentclass[letterpaper, 10 pt, conference]{ieeeconf}  
\usepackage{amsmath,graphicx}

\IEEEoverridecommandlockouts                              

\usepackage{graphics} 
\usepackage{epsfig} 
\usepackage{mathptmx} 
\usepackage{times} 
\usepackage{amsmath} 
\usepackage{amssymb}  

\title{\LARGE \bf
A Critical Analysis of the Limitation of Deep Learning based 3D Dental Mesh Segmentation Methods in Segmenting Partial Scans
}

\author{Ananya Jana$^{1}$ and Aniruddha Maiti$^{2}$ and Dimitris N. Metaxas$^{1}$
\thanks{$^{1}$Ananya Jana, Dimitris N. Metaxas are with the Department of Computer Science, Rutgers University
        {\tt\small aj611@rutgers.edu}}%
\thanks{$^{2}$Independent Contributor}
}

\begin{document}

\maketitle
\thispagestyle{empty}
\pagestyle{empty}

\begin{abstract}

Tooth segmentation from intraoral scans is a crucial part of digital dentistry. Many Deep Learning based tooth segmentation algorithms have been developed for this task. In most of the cases, high accuracy has been achieved, although, most of the available tooth segmentation techniques make an implicit restrictive assumption of full jaw model and they report accuracy based on full jaw models. Medically, however, in certain cases, full jaw tooth scan is not required or may not be available. Given this practical issue, it is important to understand the robustness of  currently available widely used Deep Learning based tooth segmentation techniques. For this purpose, we applied available segmentation techniques on  partial intraoral scans and we discovered that the available deep Learning techniques under-perform drastically. The analysis and comparison presented in this work would help us in understanding the severity of the problem and allow us to develop robust tooth segmentation technique without strong assumption of full jaw model.
\newline

\indent \textit{Clinical relevance}— Deep learning based tooth mesh segmentation algorithms have achieved high accuracy.  In the clinical setting, robustness of deep learning based methods is of utmost importance.  We discovered that the high performing tooth segmentation methods under-perform when segmenting partial intraoral scans. In our current work, we conduct extensive experiments to show the extent of this problem. We also discuss why adding partial scans to the training data of the tooth segmentation models is non-trivial. An in-depth understanding of this problem can help in developing robust tooth segmentation tenichniques.
\end{abstract}

\section{INTRODUCTION}

Intraoral scanners (IOS) are being adopted to capture digital dental impressions. 
The intraoral scanners can reconstruct the tooth surface in 3D. 
Tooth segmentation from intraoral scans is a key step in computer-aided dentistry.
\begin{figure*}[htbp]
\centering
\includegraphics[width=1.0\textwidth]{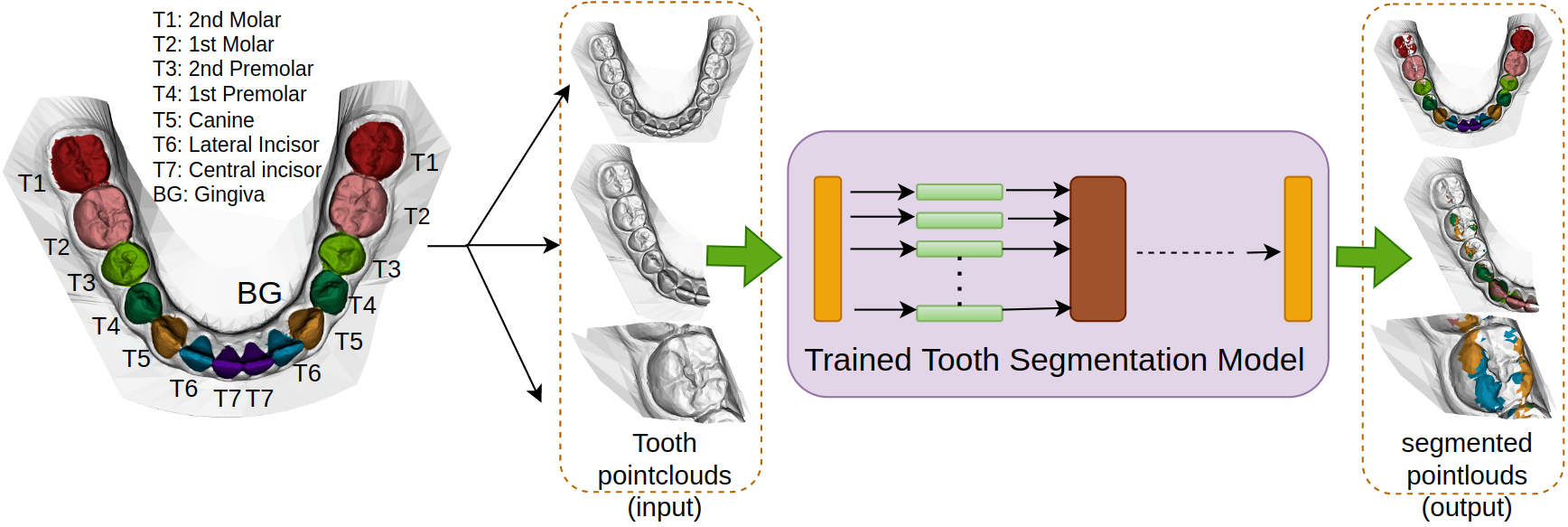}
\caption{The trained tooth segmentation algorithms fail to achieve good performance when segmenting the partial scans}
\label{fig:compareSingleTooth1}
\end{figure*}
While tooth segmentation is a first step in digital dentistry, 3D intraoral scans dataset availability is rather limited in the public domain. Due to this factor, a majority of the 3D teeth segmentation algorithms developed or analyzed so far are based on private datasets \cite{xu20183d}, 
\cite{yuan2010single}, \cite{wu2014tooth}, \cite{jana2023automatic}. There are Deep Learning methods that have been proposed for tooth segmentation from 3D intraoral scans \cite{lian2020deep}, \cite{zhao20213d}, \cite{li2022multi}, \cite{zhang2021tsgcnet}, \cite{jana20233d} Some of these algorithms have achieved excellent accuracy.  

A majority of these algorithms are built on private dataset where we do not have the information regarding subject distribution, and also these tooth segmentation algorithms are focused on the entire jaw(either lower or upper). Public dataset was not available till very recently to that extent. Recently a large public dataset has been released and hence we investigate the effectiveness of the existing methods. We are interested in testing the robustness of these methods with the help of the recently introduced public dataset 3D Teeth Seg dataset\cite{ben2022teeth3ds} for tooth segmentation. In orthodontic processes, imaging the entire jaw might not always be necessary e.g. a dental caries between the first premolar teeth and the second premolar teeth may not need the subjects's incisor teeth to be imaged or a gum recession in the gum near the left canine teeth may not require imaging of the right side molar teeth. 
Unlike other kinds of imaging modality in medical domain e.g. a CT or MRI image where the entire abdominal region is grabbed via the image taking method or device, the intraoral scans are a little different - it needs to be held manually near every tooth facing three different positions separately - buccal, occlusal and lingual side to get the 3D surface of the teeth reconstructed. This means the intraoral scans need to be maunally moved around each and every tooth to image it. Hence capturing the entire jaw might be redundant and might waste manual effort in certain orthodontic/medical treatment/diagnosis cases. The problem related to the strong assumption regarding the availability of full jaw scan for tooth segmentation can be found on literature. However, to which extend this strong assumption deteriorate the segmentation result for partial scans have not been studied extensively using all the available deep learning based segmentation methods. 

One might ask here if this shortcoming can be addressed by training the network on various partial scans. Such a training is non-trivial and the non-triviality can be attributed to another limitation in the current tooth mesh segmentation algorithms. A majority of the deep learning based  state-of-the-art tooth mesh segmentation algorithms operate on tooth mesh/dental models containing a fixed number of mesh cells. Once a partial scan is created from an intraoral scan, the number of mesh cells in the resulting scan is lower than the original intraoral scans,  
making the partial scan unfit to be used in the training. As an example we can think of a dental model containing 16k mesh cells and a partial scan containing 3k mesh cells. If the whole dental model is downsized to 3k points, the individual teeth on the dental model would lose their curvature, topological and morphological information. Resizing the partial scan to match the full intraoral scan is non-trivial due to the requirement for triangulation. The tooth mesh segmentation algorithms generally utilize k nearest neighbor methods to understand the local geometry where k represents a fixed number. When multi resolution inputs are introduced in the neural network, the same fixed k might not be sufficient to capture local geometry of the upsampled partial scan and the entire intraoral scan. In fact, working with multiple resolution of tooth mesh data is a challenge and the current state-of-the-art tooth mesh segmentation algorithms do not deal with multiple resolution of the intraoral scans.
Resizing such smaller partial meshes   that the tooth mesh contain a fixed number of mesh cells in the dental model. The main objective of this work is to investigate and compare the results. It is with this thought that we investigate the robustness of the existing state-of-the-art 3D tooth segmentation methods like - MeshSegNet \cite{lian2020deep}, TSGCNet \cite{zhang2021tsgcnet}, MBESegNet \cite{li2022multi}, GAC \cite{zhao20213d} and some of the latest generic point cloud segmentation methods such as PointNet \cite{qi2017pointnet}, PointNet++ \cite{qi2017pointnet++}, DGCNN \cite{wang2019dynamic}, BAAFNet  \cite{qiu2021semantic}, PointMLP \cite{ma2022rethinking} and  PCT \cite{guo2021pct}.  The choice of adding generic point cloud segmentation methods is due to an observation that we made - a majority of the tooth segmentation algorithms compared the tooth segmentation performance of their methods alongside the generic point cloud segmentation methods. 
 In brief, our contribution is: 
\begin{itemize}
    \item We explore the robustness of the state-of-the-art 3D tooth segmentation methods and thoroughly investigate a limitation that is severely detrimental for the deployment of these algorithms in the orthodontic/medical domain. We perform this investigation with ten different methods.
    \item We demonstrate that as we make partial intraoral scans, the methods which claim excellent accuracy for tooth-segmentation under-perform drastically.
    \item the state-of-the-art tooth segmentation algorithms are developed on mostly private datasets. We evaluate the robustness of these algorithms on the recently introduced publicly available dataset 3D Teeth Seg dataset\cite{ben2022teeth3ds}.
\end{itemize}


\section{RELATED WORKS}
Our paper is related to two broad areas of research: (a) Tooth Segmentation from Intraoral Scans and (b) 3D Shape Segmentation. 
\subsection{Tooth Segmentation from Intraoral Scans}
The conventional methods of 3D tooth segmentation usually rely upon specific prior knowledge or hand-crafted features for Tooth segmentation or tooth labeling - e.g. Kumar et. al\cite{kumar2011improved} leverage the knowledge of mesh curvature information for tooth segmentation. 
Recently, a number of deep learning-based methods have been proposed for tooth segmentation. 
In the work by Tian et al\cite{tian2019automatic}, the preprocessing of surface data was done with a sparse octree partitioning and then this preprocessed data was fed to 3D CNNs for hierarchical labeling of individual teeth. While a majority of these 3D tooth segmentation methods employ fully supervised learning, there are a few methods which employ weakly supervised or semi-supervised learning for tooth segmentation - e.g. DArch\cite{qiu2022darch}. 
The use of the tooth centroids may not make TSegNet and DArch fully effective for partial scan segmentation of partial tooth segmentation. 
\subsection{3D Shape Segmentation}
3D shape segmentation is a fundamental task in 3D computer vision. 
Deep Learning based 3D shape segmentation methods are applied on different types of input - e.g. Voxelization of surfaces as volumes to construct 3D CNNs \cite{wu20153d}, 2D rendering of 3d images \cite{su2015multi}
or on raw 3D surfaces\cite{qi2017pointnet}, \cite{qi2017pointnet++}. There are several methods that treat the task of point cloud segmentation as graph learning 
\cite{wang2018local}, \cite{wang2019graph}. Point cloud segmentation depends on the knowledge of both global and local information. Methods like PointNet\cite{qi2017pointnet} enable the network to have a global understanding of the point cloud surface. Recently there has been an interest in developing methods that are more focused on learning the local information well. An example of such methods is CurveNet\cite{xiang2021walk}.


\section{METHODS}
\subsection{Data Pre-processing}
The dataset consists of 600 subjects intraoral scans. Each raw intraoral scan consists of more than 100000 mesh cells. The intraoral scans had labels for every point. The 3D Tooth segmentation methods under discussion utilize the mesh surface features as well. Keeping this in mind, we interpolate the per-point labels to per mesh triangle label using the k-nearest neighbor method.
To reduce computation, we downsample the raw meshes to meshes containing 16000 mesh cells using quadric downsampling method. 
Each mesh cell can be characterized with four points, namely, three vertices of the mesh triangle and the center/barycenter of the mesh triangle alongwith the normals at these four points.
In our experiments we utilize the subjects who did not have wisdom teeth and hence had with teeth count $\leq$ 14.
The intraoral scans have a base which is not part of the gum. In our current work, we crop the scan to remove a the portion of base of the dental model. 
\subsection{Data Augmentation}
To improve the models' generalization ability, the training and validation sets of the data split are augmented by combining 1) random rotation, 2) random translation, and
3) random rescaling (e.g., zooming in/out) of each 3D dental surface in reasonable ranges. Specifically, along each of the three axes in the 3D space, a training/validation surface has 50
\begin{figure*}[htbp]
\includegraphics[width=1.0\textwidth]{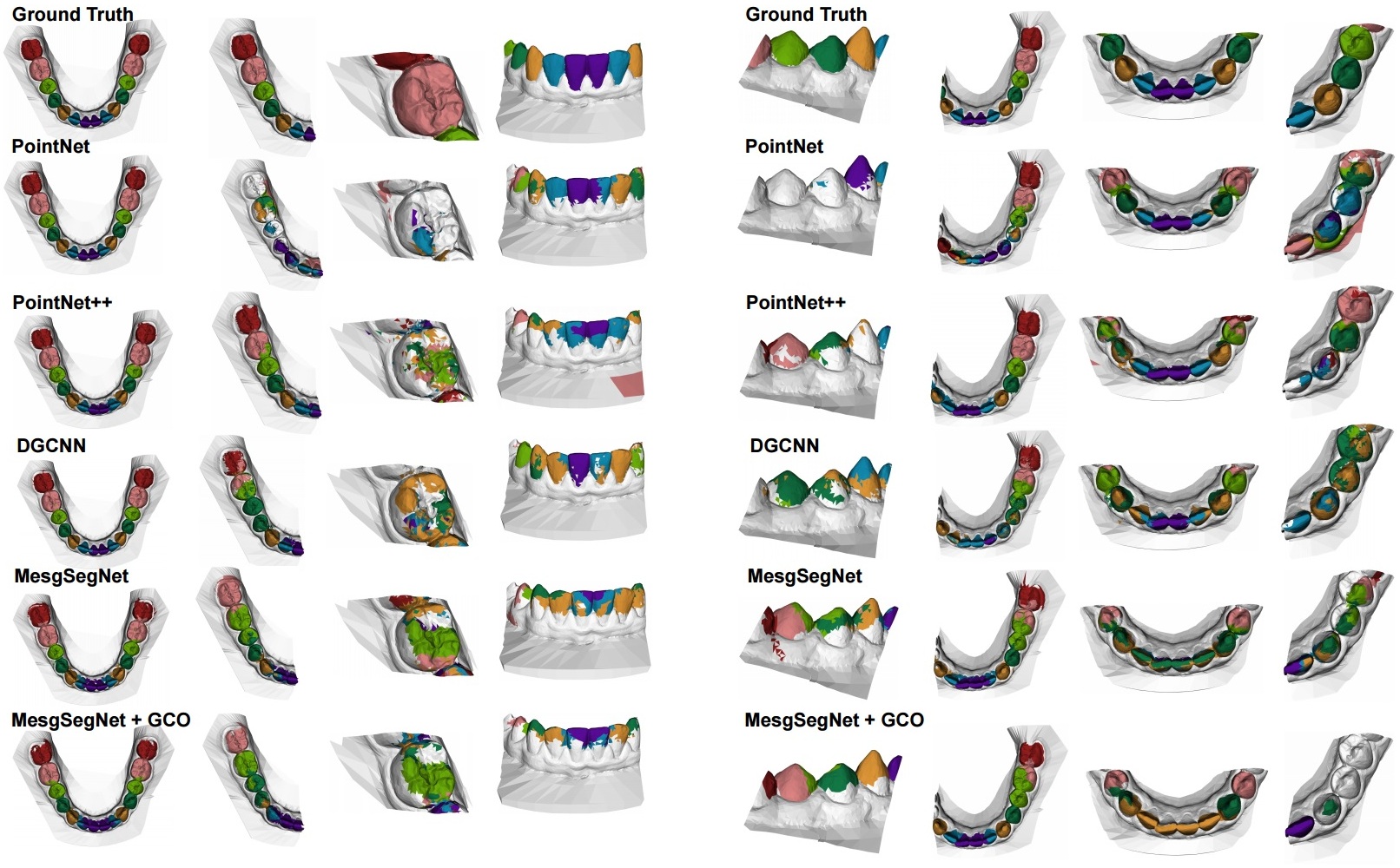}
\includegraphics[width=1.0\textwidth]{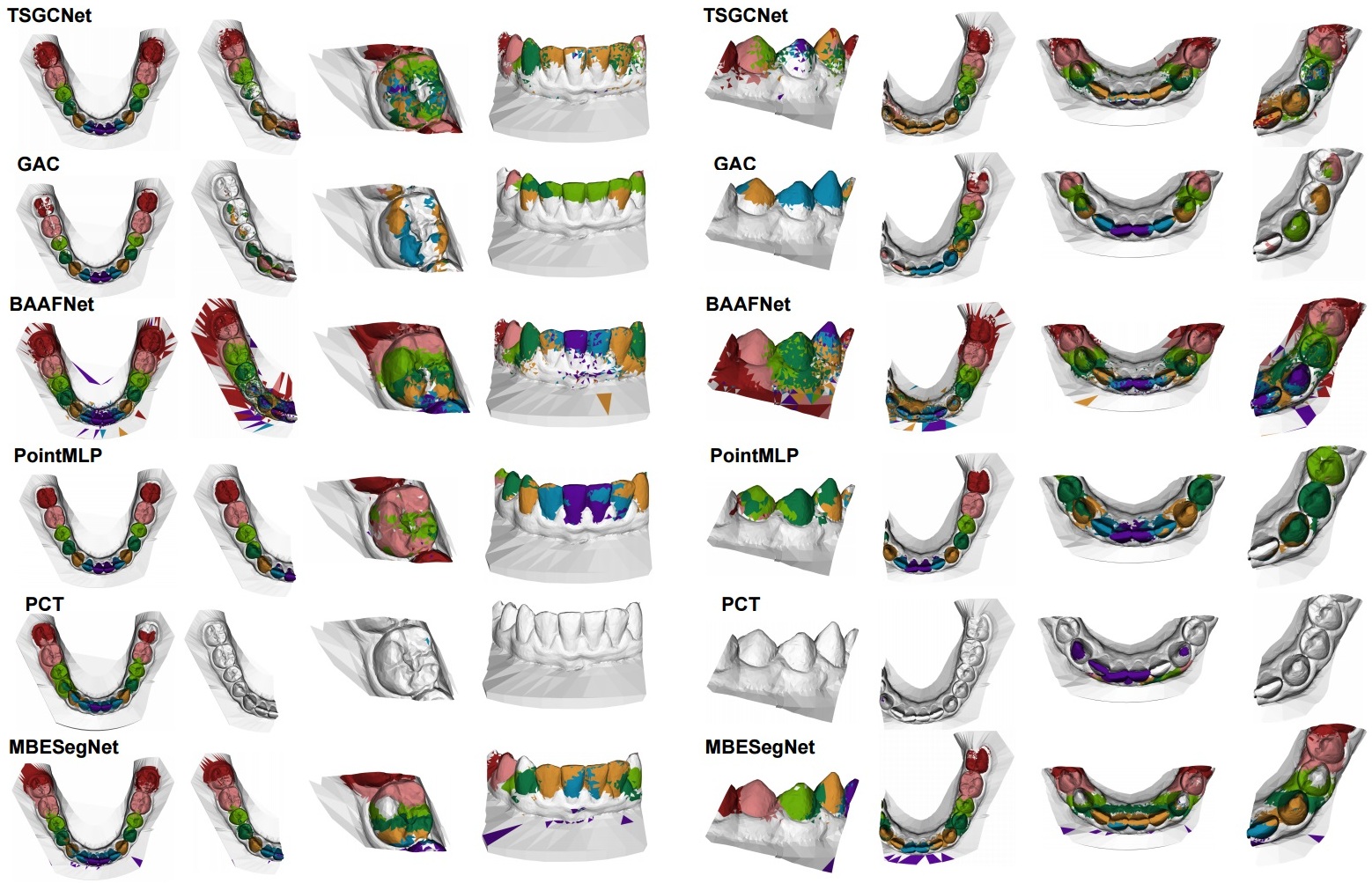}
\caption{\small  The qualitative comparison of tooth labeling for partial scans via trained methods. the leftmost column show the segmentation results of the entire scan whereas other columns show the segmentation results on partial scans. As we can see the performance deteriorates significantly for the partial scans. Whole Jaw, Half Jaw, Single Tooth, Front, Four Teeth, Ten Teeth, Eight Teeth and Three Teeth  are shown in the columns 1, 2, 3, 4, 5, 6, 7 and 8 respectively. Best viewed in color and when zoomed in.}
\label{fig:compareSingleTooth1}
\end{figure*}
\section{EXPERIMENTS}
\subsection{Dataset} The 3D Teeth Seg Challenge 2022 dataset\cite{ben2022teeth3ds} has been used in our experiments. 
In our experiments, we use the lower jaw scan of the subjects.
The task is tooth segmentation from the 3D dental model as C = 8 different semantic parts, indicating the central incisor (T7), lateral incisor (T6), canine/cuspid (T5), 1st premolar (T4), 2nd premolar (T3), 1st molar (T2), 2nd molar (T1), and background/gingiva (BG). 
We utilize 589 subjects for our experiments with  376 subjects in the training set, 95 subjects in the validation set and 118 subjects in the test set. We augment the training set and validation set data. 
\subsection{Experiment Settings}
We train ten different models on our dataset - MeshSegNet \cite{lian2020deep}, TSGCNet \cite{zhang2021tsgcnet}, MBESegNet \cite{li2022multi}, GAC \cite{zhao20213d}, PointNet \cite{qi2017pointnet}, PointNet++ \cite{qi2017pointnet++}, DGCNN \cite{wang2019dynamic}, BAAFNet  \cite{qiu2021semantic}, PointMLP \cite{ma2022rethinking} and  PCT \cite{guo2021pct}.
\subsubsection{Experiment Setting 1 (Entire Scan or Whole Jaw)} In this experimental setting, we let the trained networks predict the segmentation of the entire lower jaw intraoral scans. 
\subsubsection{Experiment Setting 2 (Half Jaw)} In this setting we crop the test set intraoral scans to have approximately half of the entire scan or half of the jaw  
by cropping a 0.48 of the intraoral scans from the right plane. 

\subsubsection{Experiment Setting 3 (Four Teeth)} In this setting, we try to roughly capture four teeth of the intraoral scans by cropping (in order): 0.56 from the left plane, then 0.15 from the bottom plane and finally 0.48 from the front plane.
\subsubsection{Experiment Setting 4 (Front)}
In this partial scan, we try to roughly capture front view of the teeth of a subject by cropping (in order): 0.30 from the right plane, 0.35 from the left plane and then 0.5 from the front plane.
\subsubsection{Experiment Setting 5 (Single Tooth)} In our experimental settings, this is the smallest partial scan that we used, roughly capturing a single tooth in deep intraoral region. The sequence of scans (in order): 0.48 from the right plane, 0.15 from the bottom plane, 0.26 from the front plane and 0.7 from the back plane. 
\subsubsection{Experiment Setting 6 (Three Teeth)} In this setting, roughly three teeth of the intraoral scans are captured by cropping (in order): 0.48 from the right plane, 0.15 from the bottom plane, 0.44 from the front plane, 0.33 from the back plane.
\subsubsection{Experiment Setting 7 (Eight Teeth)} In this setting we try to roughly capture eight teeth from the intraoral scans. The sequence of crops (in order): 0.15 from the bottom plane, 0.59 from the front plane.
\subsubsection{Experiment Setting 8 (Ten Teeth)} In this setting we try to roughly capture ten teeth from the intraoral scans by cropping 0.32 from the left plane.

\subsection{Metric} We utilize four different metrics to evaluate the performance of the tooth segmentation methods. These metrics are Overall Accuracy(OA), Dice Score (DSC), Sensitivity (SEN) and Positive Predictive Value(PPV). For all the metrics, we take an average over all the 8  class labels.

\subsection{Training Details} The ten segmentation methods have been trained for 400 epochs on RTX 8000 systems. The model yielding the best validation Dice score has been selected.

\begin{table*}[htbp]
\centering
\caption {The tooth segmentation results from ten different methods in terms of the Overall Accuracy (OA),  the Dice Score (DSC), Sensitivity (SEN) and Postive Predictive Value (PPV). Whole Jaw, Half Jaw, Front, Four Teeth, Single Tooth, Three Teeth, Eight Teeth, Ten Teeth denote the Experiment Setting 1, 2, 3, 4, 5, 6, 7 and 8 respectively. As we can see, all the ten methods perform the worst for the smallest partial scan in our experiments i.e. Single Tooth. MeshSegNet with GCO and pointMLP outperforms other methods in terms of OA, DSC and SEN. Surprisingly PCT performs best in terms of PPV. The Graph Cut Postprocessing allows for better results for MeshSegNet. But if we consider purely Deep Learning based methods, pointMLP performs comparatively robustly in segmenting partial scans.}
\begin{tabular}{|l|c|c|c|c | c | c | c | c | c | c |}
\hline
Method & Exp &  OA & DSC & SEN &  PPV & Exp &  OA & DSC & SEN &  PPV \\ \hline
{PointNet}     & Whole Jaw & 0.9167 & 0.8935 & 0.9033 & 0.9020 & Ten Teeth &  .6600 & .4973 & .5245 & .5158 \\
 & Half Jaw  & 0.4428 & 0.2210 & 0.2757 & 0.2976 & Three Teeth  & .3918 & .2395 & .3507 & .3991\\
 & Front & 0.5714 & 0.3612 & 0.5097 & 0.4142 & Eight Teeth & .6005 & .4540 & .5816 & .5073 \\
{[CVPR'17]} & Four Teeth & 0.4413 & 0.2477 & 0.4157 & 0.2935  & Single Tooth & .2824 & .1810 & .7013 & .2763 \\
\hline
{PointNet++ } & Whole Jaw & 0.8820 & 0.8432 & 0.8546 & 0.8553 & Ten Teeth & .8397 & .7944 & .7963 & .8206\\ 
& Half Jaw & 0.7909 & 0.6906 & 0.7148 & 0.7498 & Three Teeth  & .4810 & .2142 & .3856 & .4785\\
& Front & 0.5892 & 0.3061 & 0.5097 & 0.3917 & Eight Teeth & .6471 & .4031 & .6268 & .4597 \\
{[NeurIPS'17]}& Four Teeth & 0.5050 & 0.2585 & 0.4423 & 0.4238 & Single Tooth & .3945 & .2844 & .7456 & .3449 \\
\hline
{DGCNN } & Whole Jaw & 0.9311 & 0.9078 & 0.9194 & \bf{0.9143} & Ten Teeth & .8032 & .7103 & .7198 & .7533 \\
& Half Jaw & 0.6614 & 0.5249 & 0.5821 & 0.6031 & Three Teeth  & .5705 & .3762 & .4916 & .6011\\
& Front &  0.6256 & 0.4351 & 0.5483 & 0.5128 & Eight Teeth & .5932 & .4266 & .5569 & .4782 \\
{[ATG'19]} & Four Teeth & 0.5682 & 0.2761 & 0.4835 & 0.3831 & Single Tooth &  .3045 & .1707 & .7106 & .3865 \\
\hline
{MeshSegNet} & Whole Jaw & 0.8377 & 0.7875 & 0.8143 & 0.7990 & Ten Teeth & .8004 & .7382 & .7632 & .7535\\
& Half Jaw & 0.6655 & 0.5340 & 0.6010 & 0.5880 & Three Teeth  & .5198 & .2630 & .4320 & .4214\\
& Front & 0.5634 & 0.3220 & 0.4919 & 0.3737 & Eight Teeth & .5526 & .3157 & .5165 & .3755\\
{[TMI'20]}& Four Teeth & 0.4871 & 0.3307 & 0.4906 & 0.3750 & Single Tooth & .3440 & .1507 & .7267 & .2056 \\
\hline
{MeshSegNet } & Whole Jaw & 0.9125 & 0.8782 & 0.9006 & 0.8723  & Ten Teeth & .8913 & .8472 & .8826 & .8385 \\
& Half Jaw & 0.8511 & 0.6698 & \bf{0.8513} & 0.6784 & Three Teeth  & \bf{.7738} & \bf{.4966} & \bf{.8625} & .5162 \\
{[TMI'20]}& Front & \bf{0.8089} & \bf{0.5921} & \bf{0.8141} & 0.5962 & Eight Teeth & \bf{.7895} & .5469 & \bf{.8164} & .5496\\
{+GCO}& Four Teeth & \bf{0.7371} & \bf{0.5722} & \bf{0.7879} & 0.5809 & Single Tooth & \bf{.7221} & .3089 & \bf{.9221} & .3218 \\
\hline
{TSGCNet} & Whole Jaw & 0.7540 & 0.6976 & 0.7537 & 0.7057 & Ten Teeth & .6340 & .4868 & .5467 & .5504\\
& Half Jaw & 0.5463 & 0.3810 & 0.4476 & 0.4477 & Three Teeth  & .4493 & .2234 & .3937 & .3642 \\
& Front &  0.3945 & 0.1498 & 0.3784 & 0.2476 & Eight Teeth & .3713 & .1538 & .4095 & .2424\\
{[CVPR'21]} & Four Teeth &  0.4269 & 0.2021 & 0.4041 & 0.2771 & Single Tooth & .3837 &  .1717 & .7472 & .2134 \\
\hline
{GAC} &  Whole Jaw &  0.8451 & 0.7994 & 0.8080 & 0.8346  & Ten Teeth & .6779 & .5292 & .5476 & .6154\\
& Half Jaw & 0.4531 & 0.1722 & 0.2154 & 0.3956 & Three Teeth  & .4321 & .2514 & .3489 & .4245\\
& Front & 0.5317 & 0.3267 & 0.4518 & 0.4823 & Eight Teeth & .5670 & .3703 & .5271 & .4650\\
{[PRL'21]} & Four Teeth & 0.4696 & 0.2853 & 0.3967 & 0.4988 & Single Tooth  & .3140 & .2078 & .7143 & .3968 \\
\hline
{BAAFNet} & Whole Jaw & 0.5910 & 0.6015 & 0.7458 & 0.5846 & Ten Teeth & .4696 & .4482 & .5470 & .4494\\
& Half Jaw & 0.3306 & 0.2986 & 0.4139 & 0.3130 & Three Teeth  & .2992 & .2266 & .4320 & .3563\\
& Front & 0.3602 & 0.2042 & 0.4140 & 0.2580 & Eight Teeth & .3454 & .2089 & .4436 & .2795\\
{ [CVPR'21]}& Four Teeth & 0.2474 & 0.1847 & 0.3905 & 0.2267 & Single Tooth & .2743 & .1457 & .7036 & .2084 \\
\hline
{pointMLP} & Whole Jaw & \bf{0.9373} & \bf{0.9105} & \bf{0.9259} & 0.9139 & Ten Teeth & \bf{.9186} & \bf{.8840}& \bf{.8977} & \bf{.8945}\\
& Half Jaw & \bf{0.8637} & \bf{0.7702} & 0.8072 & 0.8159 & Three Teeth  & .5438 & .3074 & .4474 & .5726\\
& Front & 0.7425 & 0.5347 & 0.6959 & 0.5899 & Eight Teeth & .7680 & \bf{.6041} & .7503 & .6375\\
{[ICLR'22]}& Four Teeth & 0.6583 & 0.4309 & 0.6026 & 0.5820 & Single Tooth & .4277 & .4660 & .7585 & .5436 \\
\hline
{PCT} & Whole Jaw & 0.6192 & 0.4694 & 0.4994 & 0.5760  & Ten Teeth &  .4522 &  .1483 & .1881 & .7584\\
& Half Jaw & 0.4338 & 0.1304 & 0.1805 & \bf{0.8998}  & Three Teeth  & .4199 &  .2563 & .3252 & \bf{.8453}\\
& Front & 0.4545 & 0.2455 & 0.3621 & \bf{0.6978} & Eight Teeth & .4707 & .3381 & .4329 & \bf{.7050}\\
 {[CVM'21]} & Four Teeth & 0.4285 & 0.2833 & 0.3380 & \bf{0.9131} & Single Tooth & .3000 & \bf{.5378} & .7092 & \bf{.7626} \\
\hline
{MBESegNet} & Whole Jaw & 0.7062 & 0.6320 & 0.7002 & 0.6344  & Ten Teeth & .6539 & .5149 &  .5893 & .5359\\
& Half Jaw & 0.7040 & 0.6121 & 0.70246 & 0.6250 & Three Teeth  & .4654 & .2781 & .4758 & .4226\\
& Front & 0.3168 & 0.1112 & 0.3340 & 0.1667 & Eight Teeth & .3340 & .1411 & .3941 & .1975\\
{ [ISBI'22]} & Four Teeth & 0.3827 & 0.2853 & 0.3967 & 0.4988 & Single Tooth & .4145 & .2021 & .7669 & .2481 \\
\hline
\end{tabular}
\label{tab:allres}
\end{table*}

\begin{table*}[htbp]
\centering
\scriptsize
\caption {The tooth segmentation results from ten different methods in terms of the label wise Dice Score.}
\begin{tabular}{|c|c|c|c|c |c |c |c |c |c|}
\hline
Method  & Exp & BG & T1 & T2 & T3 &  T4 & T5 & T6 & T7  \\ \hline
{PointNet [CVPR'17]} & Whole Jaw & 0.9374 & 0.7836 & 0.9100 & 0.8853 & 0.9151 & 0.8937 & 0.8994 & 0.9236\\
& Half Jaw & 0.6949 & 0.3001 & 0.0856 & 0.0106 & 0.0013 & 0.1285 & 0.3608 & 0.1869 \\
& Front & 0.8565 & 0.5538 & 0.0119 & 0.0075 & 0.2955 & 0.2114 & 0.3845 & 0.5684 \\
& Four Teeth & 0.7556 & 0.3311 & 0.0527 & 0.1489 & 0.4544 & 0.0323 & 0.1915 & 0.0155\\
& Single Tooth & 0.5018 & 0.2447 & 0.0513 & 0.0612 & 0.4153 & 0.0375 & 0.0186 & 0.1175\\
& Ten Teeth & 0.9001& 0.6773& 0.5932& 0.4678& 0.4171& 0.1846& 0.1959& 0.5426\\
& Three Teeth & 0.5912& 0.8480& 0.0058& 0.0396& 0.0413& 0.1005& 0.1996& 0.0905 \\
& Eight Teeth & 0.8731& 0.9324& 0.0061& 0.0039& 0.0266& 0.3682& 0.6243& 0.7979\\
\hline
{PointNet++ [NeurIPS'17]} & Whole Jaw & 0.9145 & 0.7706 & 0.8931 & 0.8663 & 0.8739 & 0.8276 & 0.7724 & 0.8275 \\ 
& Half Jaw & 0.8918 & 0.6628 & 0.7218 & 0.7175 & 0.7621 & 0.6354 & 0.5899 & 0.5434\\
& Front & 0.8430 & 0.1461 & 0.0161 & 0.0215 & 0.2079 & 0.4587 & 0.3825 & 0.3732 \\
& Four Teeth & 0.7541 & 0.1386 & 0.0169 & 0.0314 & 0.3917 & 0.3512 & 0.2645 & 0.1198 \\
& Single Tooth & 0.5953 & 0.1042 & 0.2333 & 0.0850 & 0.0718 & 0.0585 & 0.1119 & 0.2617\\
& Ten Teeth & 0.9027& 0.8082& 0.8356& 0.7409& 0.7551& 0.7481& 0.7521& 0.8127 \\
& Three Teeth & 0.7173& 0.1282& 0.0817& 0.1170& 0.2678& 0.1385& 0.0974& 0.1659\\
& Eight Teeth & 0.8542& 0.2036& 0.0119& 0.0199& 0.2083& 0.6043& 0.6130& 0.7102\\
\hline
{DGCNN [ATG'19]}  & Whole Jaw & 0.9536 & 0.8205 & 0.9161 & 0.9033 & 0.9334 & 0.9258 & 0.8982 & 0.9117\\
& Half Jaw & 0.8903 & 0.5387 & 0.2997 & 0.0460 & 0.4597 & 0.6925 & 0.6325 & 0.6408 \\
& Front & 0.8969 & 0.8702 & 0.0219 & 0.0410 & 0.4556 & 0.4430 & 0.3871 & 0.3653\\
& Four Teeth & 0.8124 & 0.0638 & 0.0352 & 0.0880 & 0.6521 & 0.2115 & 0.2805 & 0.0660\\
& Single Tooth & 0.6408 & 0.3749 & 0.0290 & 0.1619 & 0.0462 & 0.0114 & 0.0377 & 0.0372\\
& Ten Teeth & 0.9260& 0.7754& 0.8208& 0.7203& 0.7363& 0.7106& 0.5394& 0.4541\\
& Three Teeth & 0.7912& 0.7998& 0.0577& 0.1144& 0.4720& 0.3665& 0.2287& 0.1800\\
& Eight Teeth & 0.9138& 0.9682& 0.0083& 0.0112& 0.0673& 0.3791& 0.3785& 0.6870\\
\hline
{MeshSegNet[TMI'20]} & Whole Jaw  &  0.9120 & 0.7026 & 0.7899 & 0.7653 & 0.8505 & 0.8211 & 0.6744 & 0.7845\\
& Half Jaw & 0.8725 & 0.2506 & 0.4421 & 0.4504 & 0.6075 & 0.6189 & 0.4336 & 0.5970 \\
& Front & 0.8430 & 0.4346 & 0.0149 & 0.0146 & 0.1152 & 0.3512 & 0.2633 & 0.5398\\
& Four Teeth &  0.7703 & 0.4506 & 0.0618 & 0.2142 & 0.2056 & 0.1878 & 0.3632 & 0.3925\\
& Single Tooth & 0.6501 & 0.0972 & 0.1325 & 0.0702 & 0.0586 & 0.0222 & 0.1509 & 0.0241 \\
& Ten Teeth & 0.8926& 0.6748& 0.7285& 0.6625& 0.7500& 0.7464& 0.6627& 0.7883\\
& Three Teeth & 0.7715& 0.2006& 0.1464& 0.2782& 0.1823& 0.2600& 0.1507& 0.1145\\
& Eight Teeth & 0.8521& 0.4701& 0.0052& 0.0110& 0.1080& 0.3529& 0.2552& 0.4715\\
\hline
{MeshSegNet[TMI'20]+GCO} & Whole Jaw & 0.9470 & 0.8408 & 0.8948 & 0.8925 & 0.916 & 0.8690 & 0.7681 & 0.8969  \\
& Half Jaw & 0.9210 & 0.6066 & 0.7563 & 0.7838 & 0.7419 & 0.2393 & 0.4848 & 0.8254\\
& Front & 0.9098 & 0.5871 & 0.6434 & 0.3430 & 0.5723 & 0.7039 & 0.3940 & 0.5836 \\
& Four Teeth & 0.8262 & 0.6036 & 0.6143 & 0.5036 & 0.5695 & 0.4263 & 0.4527 & 0.5816\\
& Single Tooth & 0.8353 & 0.2749 & 0.1124 & 0.1898 & 0.1003 & 0.3172 & 0.1377 & 0.5043 \\
& Ten Teeth & 0.9356& 0.7997& 0.8821& 0.8547& 0.8525& 0.8320& 0.7496& 0.8716\\
& Three Teeth & 0.8551& 0.5217& 0.6406& 0.5907& 0.2839& 0.3837& 0.1956& 0.5017\\
& Eight Teeth & 0.8997& 0.5774& 0.6858& 0.1945& 0.5623& 0.6567& 0.3085& 0.4905\\
\hline
{TSGCNet [CVPR'21]} & Whole Jaw  & 0.8418 & 0.5020 & 0.7378 & 0.7034 & 0.7714 & 0.7404 & 0.5850 & 0.6993\\
& Half Jaw & 0.7833 & 0.2743 & 0.3684 & 0.2354 & 0.2920 & 0.5314 & 0.3690 & 0.1949 \\
& Front & 0.7604 & 0.0017 & 0.0012 & 0.0024 & 0.0465 & 0.1961 & 0.1018 & 0.0890 \\
& Four Teeth & 0.7864 & 0.0024 & 0.0091 & 0.0711 & 0.2902 & 0.2702 & 0.1233 & 0.0645 \\
& Single Tooth & 0.7280 & 0.2073 & 0.2716 & 0.0562 & 0.0128 & 0.0157 & 0.0224 & 0.0602\\
& Ten Teeth & 0.8377& 0.4798& 0.5933& 0.5330& 0.4846& 0.4693& 0.3300& 0.1670\\
& Three Teeth & 0.7292& 0.1250& 0.1068& 0.2954& 0.2822& 0.1822& 0.0230& 0.0439\\
& Eight Teeth & 0.7488& 0.0014& 0.0008& 0.0015& 0.0063& 0.1703& 0.0900& 0.2117\\
\hline
{GAC [PRL'21]}  & Whole Jaw  & 0.8995 & 0.6330 & 0.8099 & 0.7495 & 0.8189 & 0.8365 & 0.8130 & 0.8356\\
& Half Jaw & 0.7105 & 0.3320 & 0.0334 & 0.0194 & 0.1363 & 0.0807 & 0.0320 & 0.0317\\
& Front & 0.8300 & 0.8407 & 0.0110 & 0.0230 & 0.1595 & 0.3731 & 0.3427 & 0.0337 \\
& Four Teeth & 0.7387 & 0.9320 & 0.0499 & 0.0776 & 0.2009 & 0.2329 & 0.0451 & 0.0055\\
& Single Tooth & 0.5731 & 0.4160 & 0.0271 & 0.1572 & 0.1893 & 0.0321 & 0.0252 & 0.2587\\
& Ten Teeth & 0.8517& 0.4861& 0.6798& 0.5423& 0.4986& 0.4029& 0.4304& 0.3422\\
& Three Teeth & 0.7233& 0.9362& 0.0295& 0.0083& 0.0345& 0.1582& 0.0341& 0.0871\\
& Eight Teeth & 0.8417& 0.7976& 0.0068& 0.0046& 0.0189& 0.3835& 0.4998& 0.4095\\
\hline
{BAAFNet [CVPR'21]} & Whole Jaw & 0.5016 & 0.4559 & 0.6676 & 0.6293 & 0.6634 & 0.6457 & 0.5767 & 0.6724\\
& Half Jaw & 0.3024 & 0.2781 & 0.4423 & 0.3618 & 0.3426 & 0.2005 & 0.1871 & 0.2741\\
& Front & 0.6546 & 0.0011 & 0.0019 & 0.0012 & 0.0186 & 0.2916 & 0.2868 & 0.3778 \\
& Four Teeth &  0.4262 & 0.0017 & 0.0074 & 0.0302 & 0.1283 & 0.1680 & 0.3713 & 0.3445\\
& Single Tooth & 0.5052 & 0.2512 & 0.3232 & 0.0267 & 0.0133 & 0.0140 & 0.0148 & 0.0181 \\
& Ten Teeth & 0.5113& 0.4250& 0.5497& 0.4547& 0.4340& 0.3589& 0.3797& 0.4724\\
& Three Teeth & 0.3579& 0.1254& 0.1022& 0.2305& 0.3555& 0.2494& 0.2035& 0.1888\\
& Eight Teeth & 0.6600& 0.0010& 0.0011& 0.0011& 0.0035& 0.1368& 0.2810& 0.5874\\
\hline
\end{tabular}
\label{tab:toothres}
\end{table*}

\begin{table*}[htbp]
\centering
\caption {Continuation of Table. 2. }
\scriptsize
\begin{tabular}{|c|c|c|c|c |c |c |c |c |c|}
\hline
Method  & Exp & BG & T1 & T2 & T3 &  T4 & T5 & T6 & T7  \\ \hline
{pointMLP [ICLR'22]}  & Whole Jaw &  0.9570 & 0.8152 & 0.9376 & 0.9173 & 0.9400 & 0.9218 & 0.8885 & 0.9074\\
& Half Jaw & 0.9375 & 0.6519 & 0.8569 & 0.7667 & 0.7137 & 0.8349 & 0.7131 & 0.6769\\
& Front & 0.9199 & 0.5272 & 0.0609 & 0.1514 & 0.6668 & 0.6373 & 0.6288 & 0.6857\\
& Four Teeth & 0.8015 & 0.2551 & 0.1781 & 0.5697 & 0.6197 & 0.5486 & 0.3204 & 0.1548\\
& Single Tooth & 0.6442 & 0.2254 & 0.2552 & 0.0881 & 0.1452 & 0.1257 & 0.8070 & 0.2547\\
& Ten Teeth & 0.9541& 0.8074& 0.9071& 0.8533& 0.8828& 0.8895& 0.8784& 0.8996\\
& Three Teeth & 0.8233& 0.1717& 0.0630& 0.2113& 0.3347& 0.3323& 0.2376& 0.2855\\
& Eight Teeth & 0.9198& 0.8465& 0.1220& 0.2034& 0.6763& 0.6404& 0.6777& 0.7476\\
\hline
{PCT [CVM'21]}  & Whole Jaw  & 0.7791 & 0.2974 & 0.5147 & 0.4496 & 0.3207 & 0.3654 & 0.4497 & 0.5788 \\
& Half Jaw & 0.6047 & 0.3421 & 0.0263 & 0.0183 & 0.0017 & 0.0188 & 0.0022 & 0.0297\\
& Front & 0.6271 & 0.8080 & 0.3768 & 0.0629 & 0.0014 & 0.0099 & 0.0038 & 0.0743\\
& Four Teeth & 0.5971 & 0.9944 & 0.5944 & 0.0392 & 0.0068 & 0.0274 & 0.0022 & 0.0055\\
&  Single Tooth & 0.4663 & 0.3154 & 0.0265 & 0.2460 & 0.8372 &  0.8286 & 0.8333 & 0.7492\\
& Ten Teeth & 0.6251& 0.3155& 0.0690& 0.0894& 0.0525& 0.0192& 0.0035& 0.0128\\
& Three Teeth & 0.5890& 0.9052& 0.2495& 0.0149& 0.0021& 0.0365& 0.0787& 0.1750\\
& Eight Teeth & 0.6398& 0.9430& 0.5955& 0.1879& 0.0010& 0.0020& 0.0172& 0.3184\\
\hline
{MBESegNet [ISBI'22]}  & Whole Jaw & 0.8089 & 0.4107 & 0.6989 & 0.6852 & 0.7295 & 0.6512 & 0.5464 & 0.5255 \\
& Half Jaw & 0.8138 & 0.3779 & 0.6864 & 0.6956 & 0.7371 & 0.6369 & 0.5482 & 0.4011\\
& Front & 0.7350 & 0.0018 & 0.0015 & 0.0011 & 0.0016 & 0.0750 & 0.0588 & 0.0155 \\
& Four Teeth & 0.7928 & 0.0027 & 0.0047 & 0.0018 & 0.0207 & 0.2313 & 0.3009 & 0.3005\\
& Single Tooth & 0.7192 & 0.3788 & 0.3737 & 0.0296 & 0.0140 & 0.0240 & 0.0445 & 0.0329\\
& Ten Teeth & 0.8391& 0.4839& 0.6522& 0.5281& 0.6146& 0.5563& 0.3419& 0.1032\\
& Three Teeth & 0.7915& 0.1303& 0.0740& 0.0980& 0.3278& 0.3492& 0.1936& 0.2611\\
& Eight Teeth & 0.7338& 0.0014& 0.0013& 0.0099& 0.0041& 0.1002& 0.1395& 0.1391\\
\hline
\end{tabular}

\label{tab:toothres2}
\end{table*}

\section{RESULTS}
The experimental results  are listed in Table.~\ref{tab:allres}, ~\ref{tab:toothres} and ~\ref{tab:toothres2}. The Table.~\ref{tab:allres} shows the overall accuracy Dice score, sensitivity and positive predictive value of the different methods averaged across all the class labels under the eight different experimental setups which have been described previously i.e. Whole Jaw, Half Jaw, Four Teeth, Front Teeth, Three Teeth, Eight Teeth, Ten Teeth and Single Tooth. As expected, in most of the cases the performance drops significantly for partial scans. Out of these ten methods, the MeshSegNet method with graph cut postprocessing and PointMLP method performs comparatively better. It is worth noting that the comparatively better performance of MeshSegNet with postprocessing can be attributed to the graph cut post processing rather than the  Deep Learning method, as can be understood by comparing with the MeshSegNet results without the graph cut post-processing. Although the results from PCT method shows high DSC score for a single tooth, it can be seen from the qualitative image that PCT actually fails to label any partial scan. The reason PCT still reports a high DSC value and PPV is that since for single tooth there are roughly two labels present in the partial scan and trivially labeling all the cells to any of the two labels makes the metric values high. For better understanding these nuances, we also present the segmented scans in Fig.~\ref{fig:compareSingleTooth1}. We also present the Figure \ref{fig:toothwise} for the visual comparison of the Dice score drop of the different methods under the different settings. Additionally, in the Table.~\ref{tab:toothres} and Table.~\ref{tab:toothres2}  we compare the performance of the ten different methods across the different class labels for each of the tooth-class. We demonstrate the qualitative results of 3D tooth segmentation via the different methods under the different experimental settings in Fig.~\ref{fig:compareSingleTooth1}.

\section{CONCLUSION}
We are aware of the assumption made by the 3D tooth segmentation methods about the full jaw model and the 3D tooth segmentation problem is treated
as a part segmentation problem where there are fixed number of parts of the jaw model. But this treatment is not generalizable as partial tooth scans are also equally valid in orthodontistry. There is no study which discusses and systematically analyzes this full jaw model assumption related limitation. Our current study closes this gap in existing literature. Verification of the robustness and understanding the extent of the limitation in the context of part segmentation is important because large number of tooth data are not full jaw data, but partial scans. Our work demonstrates the extent to which the performance of the trained segmentation methods deteriorates with varying sized partial scans which are medically valid.
\section{COMPLIANCE WITH EDTHICAL STANDARDS}
\label{sec:ethics}
Ethical approval was not required as confirmed by the license attached with the open access data\cite{ben2022teeth3ds}.




\bibliographystyle{unsrt}
\bibliography{refs} 
\end{document}